\pdfoutput=1
\documentclass[11pt]{article}

\usepackage[]{acl}

\usepackage{times}
\usepackage{latexsym}

\usepackage[T1]{fontenc}
\usepackage[utf8]{inputenc}

\usepackage{microtype}

\usepackage{microtype}

\usepackage{svg}
\usepackage{adjustbox}
\usepackage{multirow}
\usepackage{makecell}

 \usepackage[font=small,labelfont=bf]{caption}
 \usepackage{subcaption}
 \usepackage{caption}
\usepackage{booktabs}
\usepackage{natbib}
\usepackage{xurl}
\usepackage{hyperref}
\usepackage{bbding}
 \usepackage{xcolor}

\definecolor{taa-color}{rgb}{189,0,109}

\title{N-Shot Learning for Augmenting Task-Oriented Dialogue State Tracking}
\author{Taha Aksu\textsuperscript{$\dagger\ddag * $,}, Zhengyuan Liu\textsuperscript{$\ddag\mathsection$}, Min-Yen Kan\textsuperscript{$\dagger$}, Nancy F. Chen\textsuperscript{$\ddag$} \\
 \textit{$\dagger$} National University of Singapore\\
 \textit{$\ddag$} Institute for Infocomm Research, A*STAR\\
  \textit{$\mathsection$} CNRS@CREATE$^1$ \\
 \texttt{*taksu@u.nus.edu}}
 
\date{}
\begin{document}
\maketitle
\addtocounter{footnote}{1}
\footnotetext{ CNRS@CREATE LTD, 1 Create Way, \#08-01 CREATE Tower, Singapore 138602}

\begin{abstract}

Augmentation of task-oriented dialogues has followed standard methods used for plain-text such as back-translation, word-level manipulation, and paraphrasing despite its richly annotated structure.
In this work, we introduce an augmentation framework that utilizes belief state annotations to match turns from various dialogues and form new synthetic dialogues in a bottom-up manner. Unlike other augmentation strategies, it operates with as few as five examples.
Our augmentation strategy yields significant improvements when both adapting a DST model to a new domain, and when adapting a language model to the DST task, on evaluations with TRADE and TOD-BERT models. 
Further analysis shows that our model performs better on seen values during training, and it is also more robust to unseen values.
We conclude that exploiting belief state annotations enhances dialogue augmentation and results in improved models in $n$-shot training scenarios.

\end{abstract}
\section{Introduction}

Task-oriented dialogue (TOD) agents are the next-generation user interface and are slated to replace browsing static websites.
However, a key bottleneck in fielding such agents practically concerns adapting to new domains with few available data. In the light of this dependency in ample amounts of annotated data, {\bf data augmentation}
% a method used for increasing the diversity and size of a dataset, 
is growing in importance~\cite{feng-etal-2021-survey}.
Most augmentation methods in natural language processing (NLP) target written forms of text --- passages, news articles, {\itshape etc.} --- which operate with word- or sentence-level permutations of the original text data, synthesizing new text~\cite{textDataAug,wei-zou-2019-eda,DBLP:journals/corr/abs-1804-09541,DBLP:conf/iclr/XieWLLNJN17,kobayashi-2018-contextual}. These methods do not exploit the structure of conversational data in its entirety. We study augmenting task-oriented dialogues, a specific form of conversational data.

\begin{figure}[t]
  \centering
    \includegraphics[width=0.45\textwidth]{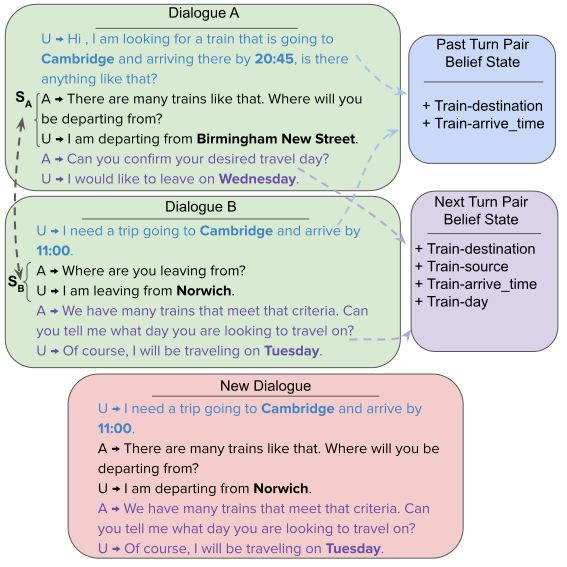}
    \caption{Scenario with two dialogues from \textit{train} booking domain. Dialogue snippets, $S_A \& S_B$, have the same dialogue function and the new dialogue created by replacing them and inserting proper slot values is still coherent end to end.}
    \label{fig:theoryExample}
\end{figure}

% A task-oriented dialogue is an information exchange where the user informs the agent of their preferences towards accomplishing a task, and the agent tries to extract them.
A TOD is a form of conversation where the aim is to accomplish a task through exchanges between a user and an agent, accounting for the user's preferences.

Within TOD, dialogue state tracking (DST) is a fundamental task, which aims to detect these preferences in a given dialogue. For this task, each pair of utterances in a dialogue is annotated with slot-label and slot-value pairs ({\itshape cf. } Figure~\ref{fig:theoryExample}: \textit{train-destination}: ``Cambridge'') and a belief state. Here, a {\it belief state} can be equated as an attribute--value store that gives the final values of each slot label (attribute) after an utterance. 

There have been several attempts to augment conversational data in the literature. \citet{Quan2019EffectiveDA} up-sample the data through word or sentence level modifications, following standard text augmentation techniques in NLP such as synonym substitution, back-translation, or paraphrasing. \citet{Kurata2016LabeledDG} perturb embeddings of single utterances and decode similarly functioned synthetic utterances. \citet{gao-etal-2020-paraphrase} create an end-to-end pipeline that finds the utterances with similar dialogue functions and trains a paraphrasing model. CoCo~\cite{li2021coco} trains a conditional user--utterance generation model, then generates synthetic turns by modifying belief states using a rule-based system and conditioning the model on the modified belief state. \citet{gritta-etal-2021-conversation} create a working graph of TOD datasets where each edge is a dialogue act and create synthetic dialogues by traversing alternative paths; however, their framework requires user acts to work with.
Critically, none of the above techniques exploit the belief state annotations of TODs within an $n$-shot scenario. 

In contrast, dialogue belief state annotations guide our approach to an effective $n$-shot augmentation method. 
%Motivate your method by giving the gist of the work and explaining it with a running example
% Our method utilizes the belief states of turn-pairs for augmentation. 
We observe that the belief state identifies the specific slots that each turn-pair discusses.  As such, belief states can be used as a proxy to represent dialogue function. 
% use the dialogue state of a turn-pair combined with the consecutive turn-pairs' dialogue states to represent its function within the dialogue. 
% Our method exploits this.  
For example, after exchanging two turn-pairs that serve the same dialogue function in separate dialogues,
coherency in both dialogues should be preserved,
if discounting necessary changes to slot values (Figure~\ref{fig:theoryExample}). Motivated by this, we delexicalize and store each turn-pair with their dialogue function to effectively

construct new dialogues from scratch.

%Tell about the experimentation setting and details
We evaluate our framework with MultiWOZ, a multi-domain dialogue dataset~\cite{budzianowski-etal-2018-multiwoz}. Each of its 10,000 dialogues is annotated with its turn belief states, system acts, and turn slots. 

We  experiment using both the previous state-of-the-art (SOTA) recurrent TRADE~\cite{wu-etal-2019-transferable} model and the transformer-based TOD-BERT~\cite{wu-etal-2020-tod} model.
% Tell about your tangible contributions and try to show the position of this study in the field
Our framework significantly increases $n$-shot performance, 

both when adapting a DST model to a new domain and when adapting a language model to the DST task. A fine-grained analysis of evaluation results reveals that models finetuned on synthetic data become robust to previously unseen slot values, and recognize seen values better.  The latter aspect accounts for the majority of the performance gain.

\section{Related Work}
\subsection{Dialogue State Tracking}
Previous DST models cumulatively keep track of utterances to obtain dialogue states \cite{Williams2007PartiallySystems,Thomson2010BayesianSystems,wang-lemon-2013-simple}. \citet{Lei2018Sequicity:Architectures} introduced Sequicity to generate belief spans as an intermediate process and improve the performance on the end task. \citet{Zhong2018Global-LocallyTracking} proposed to use a unique module for each slot, which improves the tracking of unseen slot values. The majority of these systems relied on an in-domain vocabulary and they were all evaluated on a single domain. \citet{Ramadan2018Large-ScaleSharing} proposed to jointly train the domain and state tracker using multiple bi-LSTMs and allowed the learned parameters to be shared across domains; whereas \citet{Rastogi2017ScalableTracking} used a multi-domain approach using bi-GRU where the dialogue states are defined as distributions over a candidate set derived from dialogue history. 

We use two base models in this paper. The first one, TRADE, was proposed by \citet{wu-etal-2019-transferable}.  It implements an encoder--decoder architecture and applies a copy mechanism that helps to overcome out of vocabulary (OOV) challenges. The second one, TOD-BERT~\cite{wu-etal-2020-tod}, is a task-oriented dialogue model following the transformer paradigm. It is pretrained using 9 TOD datasets with a contrastive objective function.
\subsection{Few-shot Dialogue State Tracking}
Many papers focus on the low-resource scenario in the DST field aiming to generate comparable results between low- and rich-resource settings. These invariably categorize into two approaches to address the low-resource challenge: (1) optimization functions aimed to exploit the smaller amounts of data, or (2) augmentation of the target data.

\paragraph{Few-shot Models and Techniques.}
Some approaches in the first class of solutions benefit from the recent transformer trend. One such study finetunes the GPT-2 model and reports $n$-shot slot-filling and intent recognition results on the SNIPS dataset~\cite{madotto2020language}. They achieve promising results compared to baselines with fewer shots. TOD-BERT reports results on four downstream tasks in the full- and low-resource settings~\cite{wu-etal-2020-tod}. Another line of research tries to address the problem without transformers. Span-ConverRT re-defines the slot-filling problem as turn-based span extraction that helps greatly in the few-shot setting~\cite{coope-etal-2020-span}. \citet{Huang2020Meta-ReinforcedSystems} use the model agnostic meta-learning (MAML) algorithm to adapt to new domains and show that it can outperform traditional methods with fewer data. Coach \cite{liu-etal-2020-coach}, on the other hand, breaks the slot-filling task into two components: a first slot entity detection task, followed by an entity type prediction task.

% \subsubsection{Data Augmentation For Few-shot Setting}
\paragraph{Data Augmentation for the Few-shot Setting.}
Other studies, like our approach, focus on augmentation to improve few-shot performance. \citet{Quan2019EffectiveDA} adopt four techniques for augmentation: synonym substitution, stop-word deletion, translation, and paraphrasing at the sentence level.~\citet{Kurata2016LabeledDG} start by pretraining a dialogue encoder--decoder, and then perturb the dialogue representations to back-decode synthetic dialogues. Another study by \citet{DBLP:journals/corr/abs-1810-00670} trains a logistic regression model on the small target data to detect the most informative $n$-grams and then find related samples from an out-of-domain corpus. \citet{DBLP:conf/aaai/YinSJCL20} propose a reinforcement learning setting, alternating learning between a generator and a state tracker to discover augmentation policies that benefit the end task. Two separate studies try to solve the OOV problem by enriching dialogue slot values with other values \cite{DBLP:journals/corr/abs-2002-09634,summerville-etal-2020-tame}.
\citet{liu-etal-2019-fast} train a TOD comprehension model using a synthetic data generator that simulates human-human dialogues. The transformations within the generation process are on the turn-level which limits the information flow to the rest of the dialogue. \citet{aksu-etal-2021-velocidapter} on the other hand take whole dialogues states into consideration during synthetic generation, however, their augmentation method requires manual annotation for each new domain.

\citet{campagna-etal-2020-zero} create an abstract dialogue model by defining domain templates through manual observations and then generates augmented data using these templates. Their model improves the zero-shot performance but requires manual work for each new domain.

Three studies use dialogue annotations during the augmentation process. PARG matches turns of a task-oriented dialogue by their dialogue state to create pairs for paraphrase generation~\cite{gao-etal-2020-paraphrase}, they then jointly train the paraphrase generator with the end task outperforming other dialogue augmentation baselines.
The low-resource setting defined by PARG is still required to be large enough to train a neural paraphrase model from scratch, thus limiting its applicability to emerging domains with little data.
Moreover, they do not model the interaction of a turn-pair with the next turn-pairs; as such a paraphrased utterance may be noisy, repeating a slot on the next turn.
\citet{gritta-etal-2021-conversation} create graph representations of dialogue datasets where each edge corresponds to a dialogue act by the user or system. They then extract alternative dialogues. However, they experiment only using full data settings. Additionally, their framework presumes the dialogue states are specific to each utterance, but for MultiWOZ (among other datasets) dialogue states harbor information from a pair of system--user utterances. Lastly, \citet{li2021coco} train a conditional user-utterance generation model on a large dataset, then generate synthetic dialogues by mutating the belief states through a rule-based system. This method is also limited as it requires enough data to train a conditional generation model, an unrealistic requirement for few-shot training.
% While or framework is applicable to others by combining pairs of turns theirs is not easily adaptable to datasets lacking those user acts.

\section{Method}
% Start by defining your hypothesis and the assumption it lies upon

Our method leverages a simple hypothesis, visualized in Figure \ref{fig:theoryExample}: that the function of a pair of turns in a dialogue can be defined by its slots, and its interactions with its previous and next turn-pairs. The example has two turn-pairs: $S_a$ from Dialogue A and $S_b$ from Dialogue B. The turn-pair belief states that precede both $S_a$ and  $S_b$ are composed of the same set of slot labels. The same holds for the belief states of turn-pairs following $S_a$ and $S_b$.

Thus $S_a$ and $S_b$ have the same function in the dialogue. We hypothesize the interchange of these pairs of turns (after changing the values according to the parent dialogue state) maintains a coherent dialogue. Our observations on the MultiWOZ dataset showed that this is true to a large extent for task-oriented dialogues because the belief state history represents the ongoing topic, and slot labels of the next turn give hints about the system acts.

\begin{figure}[t]
  \centering
    \includegraphics[width=0.5\textwidth]{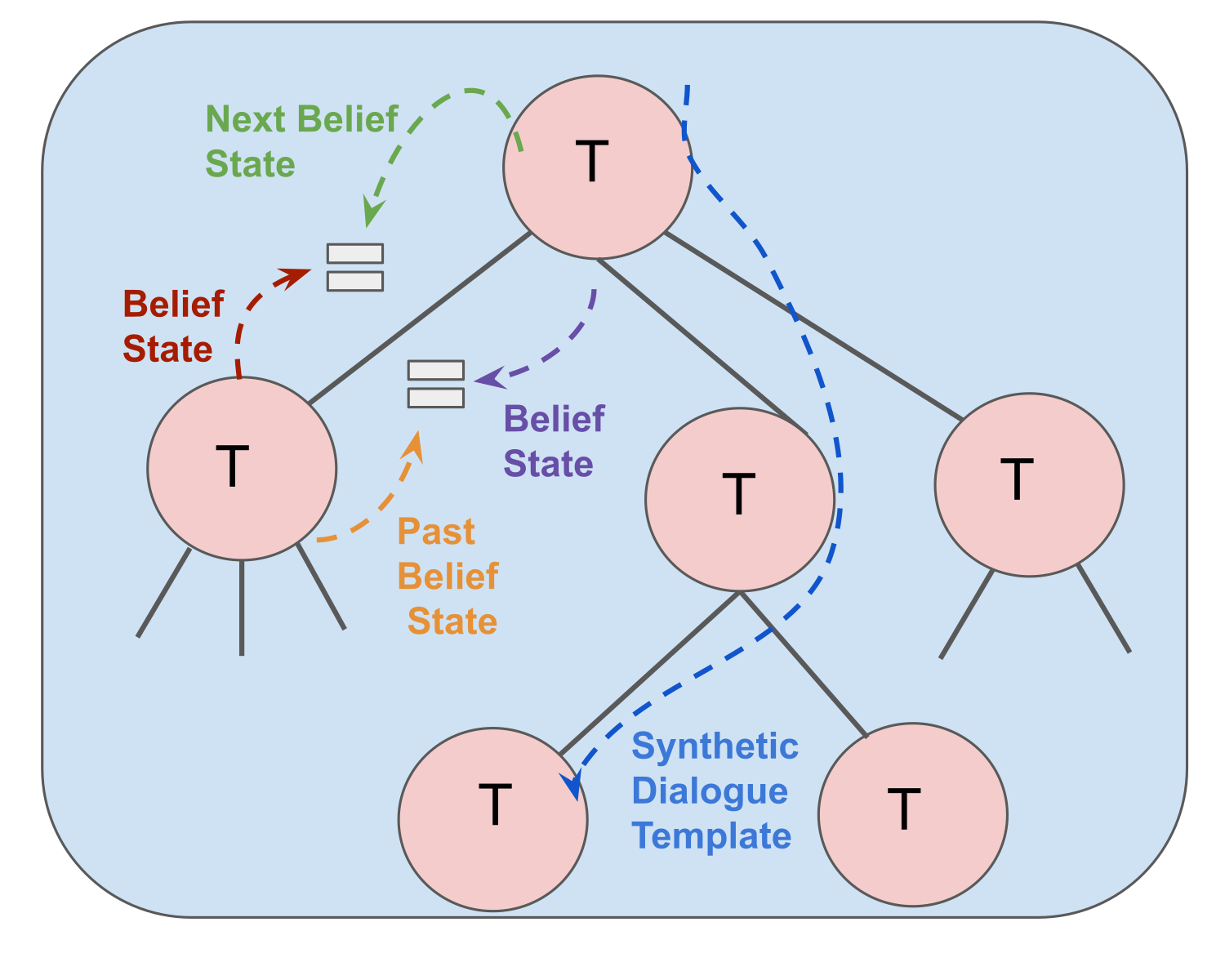}
    \caption{In our framework, dialogue templates are generated through adding proper turn-pair templates in a chain structure. The chains form a tree, which covers every possible dialogue template as a path from root to a leaf node.}
    \label{fig:tree}
\end{figure}

Our framework implements this hypothesis in three steps. In Step~1 (\S~\ref{ss:ttg}), we create turn-pair templates by delexicalizing each pair (replacing slot values with their respective slot label),

then storing each template with the previous, current, and next pair's belief states ({\it cf.} Figure~\ref{fig:template}). We also mine a dictionary of possible slot label--value pairs to be used in filling generated templates. In Step~2 (\S~\ref{ss:dtg}) we create dialogue templates by combining these pairs constrained such that two consecutive pairs' dialogue functions do not break coherency. We do this combination in a breadth-first manner, best visualized as a tree where each node is a turn-pair template, and every string of nodes from root to leaf is a dialogue template ({\it cf.} Figure~\ref{fig:tree}).
Finally in Step~3 (\S~\ref{ss:sr}), we create final synthetic dialogues by filling the slot labels in the dialogue templates ({\it cf.} Figure~\ref{fig:surface}) using the mined dictionary.

\subsection{Step 1: Turn-pair Template Generation}
\label{ss:ttg}
\begin{figure}[t]
  \centering
    \includegraphics[width=0.45\textwidth]{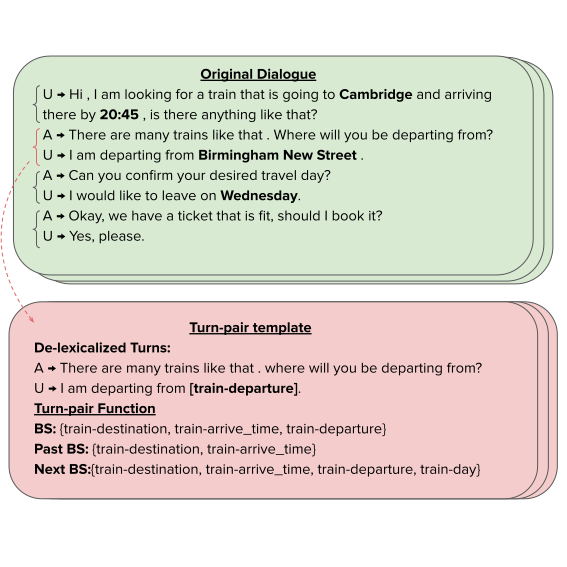}
    \caption{Sample turn-pair template (bottom, pink) and the original dialogue it is extracted from (top, green). The subject template is composed of four elements: 1) delexicalized turn utterances, and the belief state of 2) current, 3) past, and 4) next turns in the original dialogue.}
    \label{fig:template}
\end{figure}

Figure \ref{fig:template} depicts a sample turn-pair template that our framework generates. Each turn-pair template in our framework consists of a pair of turns: a system turn and a user turn. Our templates consist of pairs of turns, simply because consecutive turns (system--user) share the same dialogue state annotation. Each turn-pair template consists of a delexicalized pair of turns and a dialogue function formed as the combination of the previous, current, and next turn belief states.

%Explain that values are not labeled in text. (categorical slots, and different slots with same values)
During delexicalization we follow \cite{hou-etal-2018-sequence} to replace each slot value with ``[slot-name]''. Since MultiWOZ~2.1 does not provide indices for slot values, we manually find each value by searching in the turn-pair. 
This brings up several problems where two slots might have the same value or where some categorical values might not show up in the text ({\itshape e.g.} \textit{hotel-internet}: \{``dontcare'', ``yes'', ``no''\}). We filter out templates with the same values for different labels and leave the values for the categorical labels the same, assuming that they are independent of changes in other values. However, unlike non-categorical ones, we are limited from enriching the values of such slot types through surface realization when we fill in our templates.
Each dialogue in MultiWOZ usually starts with a salutation and ends with a farewell. To distinguish these starting--ending pairs, we define two exception cases: (1) If a template's turn-pair comes from the beginning of a dialogue, we set its previous belief state as {\it null} (start state), (2) if it comes from the ending of a dialogue we set its next belief state as {\it null} (end state). We use these two cases later in template generation to generate coherent dialogues from start to end.

\subsection{Step 2: Dialogue Template Generation}
\label{ss:dtg}
% Explain dialogue template generation
We generate each dialogue template by combining a set of turn-pair templates.  
We form our dialogue templates using a tree structure where each node corresponds to a turn-pair template, and a chain of nodes starting from a root and ending with a leaf is a dialogue template (Figure~\ref{fig:tree}). We start by defining a root node and setting its belief state as {\it null}. Initially, we ignore the next belief state condition and add every template whose previous belief state is {\it null} --- such turns are legitimate conversation starters (roots).  At each level, we mark every newly-added node as an active node. Then after each level, we iterate through active nodes and expand each node with the set of eligible templates. 
Two conditions need to be met to append Template~B to the tail of Template~A: (1) B's belief state slots should be met by A's next belief state slots and (2) A's belief state slots should be met by B's past belief state slots.
We continue adding templates until there are no active nodes. Eventually, we end up with a tree structure where each connected node represents a turn-pair and each path from the root to a leaf node is a unique dialogue template. We discard paths whose leaf nodes do not have {\it null} as the next belief state. This ensures that the dialogue template has a valid ending.

\begin{figure}[t]
  \centering
  \includegraphics[width=0.5\textwidth]{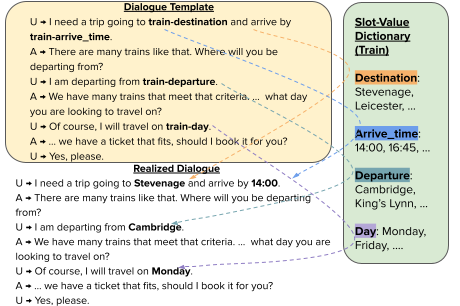}
    % \includesvg[width=0.5\textwidth]{Figures/SVGS/SR_new.svg}
    \caption{The last step in our framework, surface realization, utilizes the dictionary of slot label and slot values obtained from the original dialogues in Step~1, populating the templates with every permutation of possible values of each slot.}
    \label{fig:surface}
\end{figure}

\subsection{Step 3: Surface Realization}
\label{ss:sr}
We now fill in the delexicalized dialogue templates. Using the slot--value dictionary extracted in Step~1, we fill each dialogue with every possible slot value combination thus effectively sourcing synthetic augmented dialogues (Figure~\ref{fig:surface}). This final step returns a set of task-oriented dialogues, suitable for training (or fine-tuning) a learning system ({\it cf.} Appendix~\ref{sec:sample_dials} for sample dialogues).

\begin{table*}[hbt!]
\begin{adjustbox}{width=1\textwidth}
\begin{tabular}{|l|cc|cc|cc|cc|cc|}
    \hline
     & \multicolumn{2}{c|}{Hotel} & \multicolumn{2}{c|}{Taxi} & \multicolumn{2}{c|}{Restaurant} & \multicolumn{2}{c|}{Attraction}& \multicolumn{2}{c|}{Train}\\
    & Joint & Slot & Joint & Slot & Joint & Slot & Joint & Slot & Joint & Slot \\
    \hline
    1. Base Model (BM) trained on other 4 domains & 0.12 & 0.64 & 0.60 & 0.73 & 0.12 & 0.54 & 0.18 & 0.54 & 0.22 & 0.49 \\
    \hline
    2. BM fine tuned with 1\% data ( 84 samples) & 0.21  & 0.76  &  0.61  & 0.75  &  0.21  & 0.77 &  0.43 & 0.74  &  0.61 & 0.91\\
    \hline
    \multicolumn{1}{|c}{\textbf{5-Shot Augmentation on Target Domain}} & \multicolumn{10}{c|}{}\\
    \hline
    3. BM fine-tuned with 5 samples &  0.12 & 0.65  & 0.59  & 0.75   & 0.12 & 0.58 & 0.25 & 0.59 & 0.25  & 0.66  \\
    \hline
    4. BM fine-tuned with augmented samples & 0.12 & \textbf{0.67*}  & 0.58  & 0.75  & \textbf{0.13} & \textbf{0.62*} & \textbf{0.26} & \textbf{0.61} & \textbf{0.31*}  & \textbf{0.77*} \\
    \hline
    \multicolumn{1}{|c}{\textbf{10-Shot Augmentation on Target Domain}} & \multicolumn{10}{c|}{}\\
    \hline
    5. BM fine-tuned with 10 samples & 0.14 & 0.68 &  0.60 & \textbf{0.76} & 0.13 & 0.63 & 0.30 & 0.63 & 0.37 & 0.81\\
    \hline
    6. BM fine-tuned with augmented samples & \textbf{0.15} & \textbf{0.69} & 0.60 & \textbf{0.76}  & \textbf{0.16*} & \textbf{0.70*} & \textbf{0.32*} & \textbf{0.66*}& \textbf{0.39}& \textbf{0.83} \\
    \hline
\end{tabular}
\end{adjustbox}
  \caption{Evaluation results of TRADE model. The first row shows the zero shot results; the second row, the finetuning with 1\% data (~80 dialogues) for comparison with $n$-shot results. Each figure is an average of 10 runs. \textbf{Bolded} numbers in each section shows the best performance within that section.  ``*'' indicates statistically significant results with 95\% confidence.}
    \label{tab:Results}
\end{table*}

\begin{table}[t]
    \small
    \centering
    \begin{tabular}[l]{|l|c|c|c|}
    \hline
     Active Slot F1 & Restaurant  &Taxi  & Hotel\\
    \hline
    \multicolumn{1}{|c}{\textbf{5-Shot}} & \multicolumn{3}{c|}{}\\
    \hline
    3'. Original & 0.16 & 0.0065 & 0.20\\
    \hline
    4'. Augmented & \textbf{0.19}* &\textbf{ 0.0078}  & \textbf{0.22}*\\
    \hline
    \multicolumn{1}{|c}{\textbf{10-Shot}} & \multicolumn{3}{c|}{}\\
    \hline
    5'. Original & 0.20  & 0.010 & 0.18\\
    \hline
    6'. Augmented & \textbf{0.22}* & \textbf{0.013}* & \textbf{0.23}*\\
    \hline
\end{tabular}
\caption{TOD-BERT evaluation results over the individual \textit{restaurant}, \textit{taxi} and \textit{hotel} domains, averaged over 10 runs.Best performance within each shot level are {\bf bolded}; statistical significance ($p \geq 95\%$) is starred.}
  \label{tab:TodBertDomain}
\end{table}
\begin{table}[t!]
    \small
    \centering
    \begin{tabular}[l]{|c|c|c|c|}
    \hline
     Active Slot F1 & 20-shot & 40-shot & 80-shot  \\
    \hline
    Original samples & 0.10 & 0.16 & 0.21\\
    \hline
    % COCO augmented samples & \textbf{0.16}* & \textbf{0.21}* & \textbf{0.24}*\\
    % \hline
    Our augmented samples & \textbf{0.16}* & \textbf{0.21}* & \textbf{0.24}*\\
    \hline
    
\end{tabular}
\caption{TOD-BERT evaluation results over all domains, averaging 10 runs. Best performance within each shot level are {\bf bolded}; statistical significance ($p \geq 95\%$) is starred.}
  \label{tab:TodBertGeneral}
\end{table}

\section{Experiments}
\subsection{Dataset, Models and Evaluation}

We conduct experiments on MultiWOZ, a well-known dataset in the DST field. When compared to its counterparts like WOZ~\cite{wen-etal-2017-network}, DSTC2~\cite{henderson-etal-2014-second} and Restaurant-8k~\cite{coope-etal-2020-span}, MultiWOZ is the richest, combining several domains with a variety of slot labels and values. MultiWOZ is a multi-domain dialogue dataset that covers 10,000 dialogues between clerks and tourists, each annotated with turn belief states, system acts, and turn slots.
Following prior works~\cite{wu-etal-2019-transferable,wu2020todbert} we conduct our experiments on 5 of 7 domains leaving \textit{hospital} and \textit{police} domains out as their validation and test sets sample quantity is very low. 

We wish to assess how fine-tuning with our augmented data affects model performance.  We experiment
with the TRADE and TOD-BERT models~\cite{wu2020todbert,wu-etal-2019-transferable} to assess whether their base performance can be improved using our augmentation framework.For both models, we follow the fine-tuning experiments done by \cite{wu-etal-2019-transferable}: we train a base model on four domains and then fine-tune this model with small sets of randomly sampled data from the remaining left-out target domain (5- or 10-shots). We compare this against the scenario where we apply our augmentation framework on the small set before fine-tuning. 

Due to space limitations, we present results only for the subset of the \textit{restaurant}, \textit{taxi}, and \textit{hotel} domains in TOD-BERT.  These three domains cover almost every unique slot in the MultiWOZ dataset, and is thus representative. 
% Min3: make it clear why you do this.
We conduct an additional experiment for TOD-BERT, training/testing with data from all domains in several few shot settings (20-, 40-, and 80-shot).
% \taa{and again report results of training with augmented data by our framework and CoCo.}

We evaluate TRADE using the metrics proposed by~\citet{wu-etal-2019-transferable}: Slot Accuracy and Joint Accuracy. {\it Slot Accuracy} measures the proportion of correctly predicted slot values; while {\it Joint Accuracy} is more coarse-grained, measuring the correctly predicted turn dialogue states. To predict a turn dialogue state correctly means that all its contained slot values are predicted correctly. Also, when a slot is not mentioned in the utterance the ground truth for that slot becomes {\it None}. This results in utterances having ground truth slot values which mostly consist of the value {\it None}. We observe that in our few-shot experiments, unlike TRADE, TOD-BERT model returns predictions consisting only of {\it None} values. We believe that the discrepancy is attributable to TRADE's copy mechanism, which the TOD-BERT model lacks.  To better assess the contribution of our augmentation approach, we use Active Slot Accuracy~\cite{dingliwal-etal-2021-shot} for the TOD-BERT experiments, which is the accuracy of slot value predictions for all non-{\it None} values.

\subsection{Implementation and Training Settings}
\label{sec:training_details}
We adjust our training settings to facilitate a fair comparison among the models trained on different data sizes (original versus augmented).
For the TRADE model, we use the default hyperparameter settings reported in the original paper. For TOD-BERT,  we change the training batch size to 4 and the evaluation batch size to 8, the development set evaluation frequency to 1 evaluation per 200 steps,  set the terminating condition to early stopping bounded by a maximum number of steps. 
For our augmented fine-tuning model training, we fine-tune the base model on synthetic data for $N$/2 steps, followed by fine-tuning on the mixture of original and synthetic data for another $N$/2 steps.  We perform this mixing of original samples in the latter part of fine-tuning to ensure that the model is exposed to a diverse set of samples, while not significantly deviating from the original distribution.  This is conceptually similar to the notion of experience replay in reinforcement learning.

\begin{figure*}[t]
  \centering
    \includegraphics[width=\textwidth]{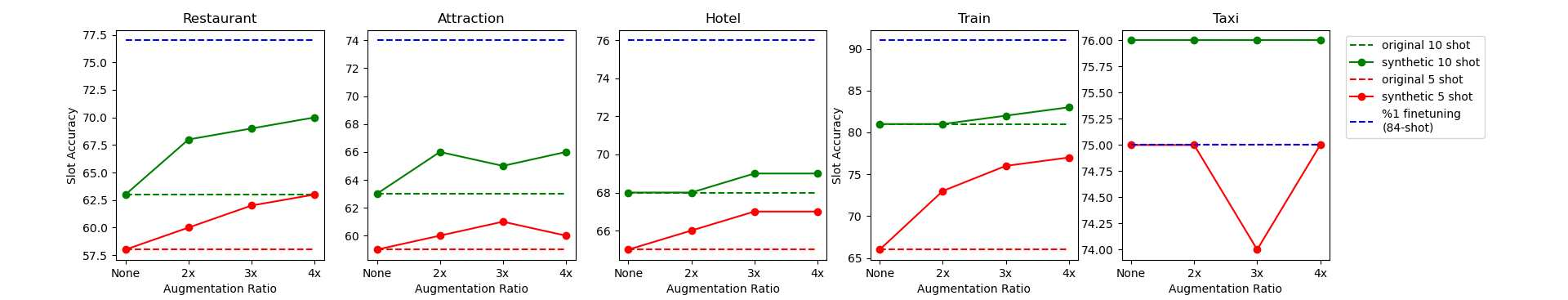}
    \caption{Effects of the augmentation ratio on TRADE model by domain. The dashed blue line represents the performance of fine-tuning with 1\% of full data ($\sim$80 dialogues) for comparison as a pseudo upper bound [Note $y$-axis scales differ per chart]. }
    \label{fig:AllRatios}
\end{figure*}
\subsection{Results}
\paragraph{TRADE Experiments (Table~\ref{tab:Results}).}
\label{TradeResults}
We report the significance of results with 95\% confidence along with averages over 10 runs. Our framework can sustain the model performance in all five domains and significantly improves over baseline (Row 1) in either the 5- (Row~4) or 10-shot (Row~6) scenarios in four of the five domains, where most results are statistically significant at the $p \geq 0.95$ level.
% Min3: add and fix BUG
These results also greatly improve over fine-tuning using just 5 or 10 target domain samples (compare Row 3 against 4, and Row 5 against 6).
Overall, applying our augmentation framework yields a macro-averaged improvement of $3.2$\% slot accuracy and $1.5\%$ joint accuracy.  As a pseudo-upper bound, we compare our method against fine-tuning over 80 shots (roughly 1\% of the target domain data, represented by Row 2), and see that our approach significantly closes this performance gap. 

The exception is the \textit{taxi} domain where the augmented data does not result in significant change. We believe this is due to \textit{taxi} domain slots having a higher variety in values than in other domain slots.  This results in many OOV values in the test set. The TRADE model thanks to its copy mechanism, adapts well to these OOV with fewer data. The fact that the performance of the base model fine-tuned with 1\% of data is already reached by fine-tuning the same model within a 5-shot scenario (compare Row~2 and Row~1's {\it taxi} column) supports our claim.

\paragraph{TOD-BERT Experiments (Tables~\ref{tab:TodBertDomain} and~\ref{tab:TodBertGeneral}).}

With TOD-BERT, we examine our framework's effect on both domain and task adaptation. 

\noindent $\cdot$ Table~\ref{tab:TodBertDomain} shows results for domain adaptation, and the figures are comparable to those in Table~1 for TRADE.  We number the rows with primes ($'$) to imply the corresponding results from the TRADE experiments.  We follow the same setting as above for TRADE (train on 4 other domains, test on target domain).  We observe uniformly improved results over the few shot fine-tuning, as we did for TRADE, proving the agnostic feature of our framework.

\noindent $\cdot$ Table~\ref{tab:TodBertGeneral} shows results for task adaptation.  Here, the TOD-BERT model has no familiarity with the DST task at all, thus fine-tuning is an adaptation to the task itself.  This is a more challenging scenario.  Again, we see uniform improvement, especially for the lower-shot scenarios (20- and 40-).  

The results for both are consistent and in favor of our framework. Our framework helps in both cases: (1) LM adaptation to a new task ({\itshape e.g.} DST), and (2) LM adaptation to a new task-oriented dialogue domain ({\itshape e.g.} \textit{restaurant}).

\subsection{How Does Augmentation Improve Performance?}
\label{TOD-BERTAnalyze}
% Divide to benefits to seen and unseen values
\begin{table}[t]
% \begin{adjustbox}{width=1\textwidth}
    \small
    \centering
    \begin{tabular}[l]{|c|c|c|}
    \hline
     Recall & Unseen Values  & Seen Values\\
    \hline
    \textbf{All-domains} & \multicolumn{2}{c|}{}\\
    \hline
    Original & 0.1 e-3  & 0.24 \\
    \hline
    Augmented & \textbf{0.2 e-3} & \textbf{0.28}\\
    \hline
    \textbf{Restaurant} & \multicolumn{2}{c|}{}\\
    \hline
    Original & 1.5 e-3 & 0.20\\
    \hline
    Augmented & \textbf{2.3 e-3} & \textbf{0.26}\\
    \hline
    \textbf{Taxi} & \multicolumn{2}{c|}{}\\
    \hline
    Original & 6.3 e-3 & 0.16\\
    \hline
    Augmented & \textbf{6.8 e-3} & \textbf{0.21}\\
    \hline
    \textbf{Hotel} & \multicolumn{2}{c|}{}\\
    \hline
    Original & 0.5 e-3 & 0.30\\
    \hline
    Augmented & \textbf{1.0 e-3} & \textbf{0.32}\\
    \hline
\end{tabular}
% \end{adjustbox}
    \caption{TOD-BERT evaluation results, subdivided between on seen and unseen values, averaged over 10 runs, with best results per section in \textbf{bold.}}
  \label{tab:SeenUnSeenValueExps}
\end{table}

\begin{table}[t]
\linespread{1.1}
    \small
    \centering
    \begin{tabular}[l]{|p{3.5cm}<{\centering}|p{1.2cm}<{\centering}|p{1.2cm}<{\centering}|}
    \hline
    Error type & Original & Synthetic \\
    \hline
    restaurant-food & 2,041 & 1,675 \\
    \hline
    restaurant-pricerange  & 1,210 & 603 \\
    \hline
    restaurant-name & 1,133 & 1,061 \\
    \hline
    restaurant-area &  853 & 480 \\
    \hline
    restaurant-book day & 743 & 335\\
    \hline
    restaurant-book people & 740 & 212\\
    \hline
    restaurant-book time & 1,119 & 347\\
    \hline
\end{tabular}
\caption{Fine-grained \textit{restaurant} domain errors, for the original and augmented TRADE model, classified by slot type.}
  \label{tab:errAnalysis}
\end{table}

\begin{table*}[t]
\begin{adjustbox}{width=1\textwidth}
\begin{tabular}{|c|cc|cc|cc|cc|cc|}
    \hline
     & \multicolumn{2}{c|}{Hotel} & \multicolumn{2}{c|}{Taxi} & \multicolumn{2}{c|}{Restaurant} & \multicolumn{2}{c|}{Attraction}& \multicolumn{2}{c|}{Train}\\
    & Joint & Slot & Joint & Slot & Joint & Slot & Joint & Slot & Joint & Slot \\
    \hline
    \textbf{5 Shot Augmentation on Target Domain} & \multicolumn{10}{c|}{}\\
    \hline
    BM fine-tuned with CoCo & 0.12 & 0.66  & \textbf{0.60}  & 0.75  & \textbf{0.13} & \textbf{0.62} & 0.24 & 0.58 & 0.27  & 0.69 \\
    \hline
    BM fine-tuned with our framework & 0.12 & \textbf{0.67}  & 0.58  & 0.75  & \textbf{0.13} & \textbf{0.62} & \textbf{0.26} & \textbf{0.61} & \textbf{0.31}  & \textbf{0.77} \\
    \hline
    \textbf{10 Shot Augmentation on Target Domain} & \multicolumn{10}{c|}{}\\
    \hline
    BM fine-tuned with CoCo & \textbf{0.15} & 0.68 & \textbf{0.61} & 0.75   & \textbf{0.16} & 0.67 & 0.31 & 0.64  & \textbf{0.39}& 0.82 \\
    \hline
    BM fine-tuned with our framework & \textbf{0.15} & \textbf{0.69} & 0.60 & \textbf{0.76}  & \textbf{0.16} & \textbf{0.70} & \textbf{0.32} & \textbf{0.66}& \textbf{0.39}& \textbf{0.83} \\
    \hline
\end{tabular}
\end{adjustbox}
  \caption{ Evaluation results of TRADE model comparing our augmentation framework to the upperbound CoCo model pre-trained on full training data (including target domain).}
    \label{tab:trade_coco}
\end{table*}

\begin{table}[t]
% \begin{adjustbox}{width=1\textwidth}
    \small
    \centering
    \begin{tabular}[l]{|c|c|c|c|}
    \hline
     Active Slot F1 & Restaurant  &Taxi  & Hotel\\
    \hline
    \textbf{5 Shot} & \multicolumn{3}{c|}{}\\
    \hline
    CoCo &  0.17 & 0.0047 & 0.21\\
    \hline
    Ours & \textbf{0.19} &\textbf{ 0.0078}  & \textbf{0.22}\\
    \hline
    \textbf{10 Shot} & \multicolumn{3}{c|}{}\\
    \hline
    CoCo &  \textbf{0.22} & 0.0114  & 0.21 \\
    \hline
    Ours & \textbf{0.22} & \textbf{0.0132} & \textbf{0.23}\\
    \hline
\end{tabular}
% \end{adjustbox}
\caption{ Evaluation results of TOD-BERT model comparing our augmentation framework to the upperbound CoCo model pre-trained on full training data (including target domain).}
  \label{tab:TodBert_coco}
\end{table}

To study the reason behind the performance gain by augmentation, we dispart our test set samples into two groups: samples with unique values that do not show up during training, and samples with values seen during training. We then evaluate the TOD-BERT model trained with original and synthetic data on these two separate groups, {\itshape cf. } Table~\ref{tab:SeenUnSeenValueExps}. The results suggest that although, augmentation increases robustness to unseen values in all domains, the largest part of the contribution is on seen values. This is expected since our framework uses the same set of values as in small original dialogue set during surface realization.

Note that for the ``All-domains'' section in the table the improvement on unseen values is smaller compared to domain-specific sections (\textit{Restaurant}, \textit{Taxi}, \textit{Hotel}), this is because, in the former, the model learns DST task from scratch thus exploiting seen-values to learn the task overweighs to generalizing over unseen values. Whereas for the latter, robustness to unseen values gets higher learning priority since the model is already familiar with the DST task from training on other 4 domains.

This analysis shows that our framework helps the model to exploit slots that have a bounded value pool with less unique values while also making it robust to unseen values for slots with broader value pools. 

\subsection{Effect of Augmentation Ratio}
We run our framework with several different augmentation ratios in both the 5 and 10 shot cases to inspect if the synthetic data amount affects the results proportionally. Figure \ref{fig:AllRatios} shows the results for the TRADE model in all 5 domains. Our framework outperforms base fine-tuning steadily, and the amount of synthetic data affects the results proportionally in every case except the \textit{taxi} domain as explained before  ({\itshape cf.} Section \ref{TradeResults}). 

\subsection{Fine-grained Error Analysis}
\subsubsection{Slot-type Errors}

Apart from performance in evaluation metrics we also analyze the error rates of the TRADE model in each specific slot type in the \textit{restaurant} domain and compare results with and without our framework. Table~\ref{tab:errAnalysis} shows the results. Our framework consistently reduces error rates in every single slot type. The drop in the error rate is least remarkable for the name and food slots, we believe this is because the challenge in these slots is most largely unknown vocabulary words. Our framework enriches the dialogue templates with values from the original set. Thus it is less helpful for those slots suffering from the unknown slot value problem and shows more significant improvements on slots with arguably more isolated vocabulary ({\itshape e.g.} Book-day: 1, 2, 3, {\itshape etc.} or price range: cheap, moderate, expensive).

To support the significance of results on fine-grained slot error types, we use McNemar's test ($\alpha = 0.01$) upon creating the confusion matrix between our framework and original fine-tuning. The results suggest that synthetic data fine-tuning shows statistically significant improvements over the original data fine-tuning, with $p < \alpha$. 

\subsection{Comparison against CoCo Model}

To better locate the position of our framework in the literature we repeat target domain experiments using another dialogue augmentation method: CoCo~\cite{li2021coco}. However, CoCo is a learning-based approach that requires rich amounts of data, so it is unfair to expect it to learn from only a few shots (5/10). 
Instead, we use the pretrained weights that are provided by the original CoCo paper and treat it as an upper bound because it is trained on the full training data (including the target domain for leave-one-out experiments) whereas our framework uses only the provided few dialogues during augmentation. 

Tables~\ref{tab:trade_coco} and \ref{tab:TodBert_coco} give the results for TRADE and TOD-BERT, respectively. Despite the advantageous standing of CoCo, our framework outperforms CoCo in all domains for the TOD-BERT model and shows either superior or comparable results on TRADE. 

\subsection{Effect of Template Generation}
\label{sec:ablation}
We conduct an ablation study to see the effect of dialogue template generation by re-running the TOD-BERT target domain experiments for \textit{hotel} and \textit{restaurant} domains with a simpler baseline, where we use only the original $n$ dialogues as templates and perform surface realization.

The results in Table~\ref{tab:TodBert_ablation} show that template generation improves results compared only surface realization in most of the cases. Our template generation strategy offers higher diversity to the samples but it might bring up noisy samples along, whereas only surface realization is less noisy but lacks the diversity that novel templates contribute.
%This might suggest that as the number of shots increases template generation addition can have diminishing returns but we leave this as an exploration for future work.

\begin{table}[t!]
% \begin{adjustbox}{width=1\textwidth}
    \small
    \centering
    \begin{tabular}[l]{|c|c|c|c|}
    \hline
     Active Slot F1 & Restaurant & Hotel\\
    \hline
    \textbf{5 Shot} & \multicolumn{2}{c|}{}\\
    \hline
    Full pipeline & \textbf{0.183} &\textbf{ 0.255}\\
    \hline
    Only SR & 0.157 & 0.250\\
    \hline
    \textbf{10 Shot} & \multicolumn{2}{c|}{}\\
    \hline
    Full pipeline & 0.198  & \textbf{0.258}\\
    \hline
    Only SR & \textbf{0.237} & 0.243\\
    \hline
\end{tabular}
% \end{adjustbox}
\caption{TOD-BERT target domain experiments comparing full pipeline (first row) against only surface realization (second row). Each number corresponds to an average of 3 runs.}
  \label{tab:TodBert_ablation}
\end{table}

\section{Conclusion}

Our framework showcases a distinct approach to dialogue augmentation, where, unlike other studies, we apply the modification not on a datum/sample level (\textit{i.e} modifying utterances or words in an utterance) but on the data level exchanging information among different samples. We apply this concept within TODs as their dialogue states are like blueprints detailing each dialogue separately which can be used to partition and reconstruct new dialogue samples from scratch. 

Experiments on MultiWOZ dataset using both the TRADE and TOD-BERT models suggest that our framework consistently improves the performance of the base-model it is applied to. This is true both when adapting the model to the DST task from scratch and also when adapting a model pretrained on DST task to a new domain. The performance boost behind our augmentation framework comes mostly from performance increase on seen values during training although it also makes the model more robust to unseen values. Showing that our framework consistently improves the few-shot performance over the DST task we believe it can open doors for many other TOD tasks in limited data scenarios.

\section{Acknowledgements}
This research was supported by the SINGA scholarship from A*STAR and by the National Research Foundation, Prime Minister’s Office, Singapore under its Campus for Research Excellence and Technological Enterprise (CREATE) programme. We would like to thank anonymous reviewers for their insightful feedback on how to improve the paper. 

\clearpage
\newpage

\bibliography{anthology,custom}

\begin{thebibliography}{37}
\expandafter\ifx\csname natexlab\endcsname\relax\def\natexlab#1{#1}\fi

\bibitem[{Aksu et~al.(2021)Aksu, Liu, Kan, and
  Chen}]{aksu-etal-2021-velocidapter}
Ibrahim~Taha Aksu, Zhengyuan Liu, Min-Yen Kan, and Nancy Chen. 2021.
\newblock \href {https://aclanthology.org/2021.sigdial-1.14} {Velocidapter:
  Task-oriented dialogue comprehension modeling pairing synthetic text
  generation with domain adaptation}.
\newblock In \emph{Proceedings of the 22nd Annual Meeting of the Special
  Interest Group on Discourse and Dialogue}, pages 133--143, Singapore and
  Online. Association for Computational Linguistics.

\bibitem[{Budzianowski et~al.(2018)Budzianowski, Wen, Tseng, Casanueva, Ultes,
  Ramadan, and Ga{\v{s}}i{\'c}}]{budzianowski-etal-2018-multiwoz}
Pawe{\l} Budzianowski, Tsung-Hsien Wen, Bo-Hsiang Tseng, I{\~n}igo Casanueva,
  Stefan Ultes, Osman Ramadan, and Milica Ga{\v{s}}i{\'c}. 2018.
\newblock \href {https://doi.org/10.18653/v1/D18-1547} {{M}ulti{WOZ} - a
  large-scale multi-domain {W}izard-of-{O}z dataset for task-oriented dialogue
  modelling}.
\newblock In \emph{Proceedings of the 2018 Conference on Empirical Methods in
  Natural Language Processing}, pages 5016--5026, Brussels, Belgium.
  Association for Computational Linguistics.

\bibitem[{Campagna et~al.(2020)Campagna, Foryciarz, Moradshahi, and
  Lam}]{campagna-etal-2020-zero}
Giovanni Campagna, Agata Foryciarz, Mehrad Moradshahi, and Monica Lam. 2020.
\newblock \href {https://doi.org/10.18653/v1/2020.acl-main.12} {Zero-shot
  transfer learning with synthesized data for multi-domain dialogue state
  tracking}.
\newblock In \emph{Proceedings of the 58th Annual Meeting of the Association
  for Computational Linguistics}, pages 122--132, Online. Association for
  Computational Linguistics.

\bibitem[{Coope et~al.(2020)Coope, Farghly, Gerz, Vuli{\'c}, and
  Henderson}]{coope-etal-2020-span}
Samuel Coope, Tyler Farghly, Daniela Gerz, Ivan Vuli{\'c}, and Matthew
  Henderson. 2020.
\newblock \href {https://doi.org/10.18653/v1/2020.acl-main.11}
  {{S}pan-{ConveRT}: {F}ew-shot span extraction for dialog with pretrained
  conversational representations}.
\newblock In \emph{Proceedings of the 58th Annual Meeting of the Association
  for Computational Linguistics}, pages 107--121, Online. Association for
  Computational Linguistics.

\bibitem[{Dingliwal et~al.(2021)Dingliwal, Gao, Agarwal, Lin, Chung, and
  Hakkani-Tur}]{dingliwal-etal-2021-shot}
Saket Dingliwal, Shuyang Gao, Sanchit Agarwal, Chien-Wei Lin, Tagyoung Chung,
  and Dilek Hakkani-Tur. 2021.
\newblock \href {https://aclanthology.org/2021.eacl-main.148} {Few shot
  dialogue state tracking using meta-learning}.
\newblock In \emph{Proceedings of the 16th Conference of the European Chapter
  of the Association for Computational Linguistics: Main Volume}, pages
  1730--1739, Online. Association for Computational Linguistics.

\bibitem[{Feng et~al.(2021)Feng, Gangal, Wei, Chandar, Vosoughi, Mitamura, and
  Hovy}]{feng-etal-2021-survey}
Steven~Y. Feng, Varun Gangal, Jason Wei, Sarath Chandar, Soroush Vosoughi,
  Teruko Mitamura, and Eduard Hovy. 2021.
\newblock \href {https://doi.org/10.18653/v1/2021.findings-acl.84} {A survey of
  data augmentation approaches for {NLP}}.
\newblock In \emph{Findings of the Association for Computational Linguistics:
  ACL-IJCNLP 2021}, pages 968--988, Online. Association for Computational
  Linguistics.

\bibitem[{Gao et~al.(2020)Gao, Zhang, Ou, and Yu}]{gao-etal-2020-paraphrase}
Silin Gao, Yichi Zhang, Zhijian Ou, and Zhou Yu. 2020.
\newblock \href {https://doi.org/10.18653/v1/2020.acl-main.60} {Paraphrase
  augmented task-oriented dialog generation}.
\newblock In \emph{Proceedings of the 58th Annual Meeting of the Association
  for Computational Linguistics}, pages 639--649, Online. Association for
  Computational Linguistics.

\bibitem[{Gritta et~al.(2021)Gritta, Lampouras, and
  Iacobacci}]{gritta-etal-2021-conversation}
Milan Gritta, Gerasimos Lampouras, and Ignacio Iacobacci. 2021.
\newblock \href {https://doi.org/10.1162/tacl_a_00352} {Conversation graph:
  Data augmentation, training, and evaluation for non-deterministic dialogue
  management}.
\newblock \emph{Transactions of the Association for Computational Linguistics},
  9:36--52.

\bibitem[{Henderson et~al.(2014)Henderson, Thomson, and
  Williams}]{henderson-etal-2014-second}
Matthew Henderson, Blaise Thomson, and Jason~D. Williams. 2014.
\newblock \href {https://doi.org/10.3115/v1/W14-4337} {The second dialog state
  tracking challenge}.
\newblock In \emph{Proceedings of the 15th Annual Meeting of the Special
  Interest Group on Discourse and Dialogue ({SIGDIAL})}, pages 263--272,
  Philadelphia, PA, U.S.A. Association for Computational Linguistics.

\bibitem[{Hou et~al.(2018)Hou, Liu, Che, and Liu}]{hou-etal-2018-sequence}
Yutai Hou, Yijia Liu, Wanxiang Che, and Ting Liu. 2018.
\newblock \href {https://aclanthology.org/C18-1105} {Sequence-to-sequence data
  augmentation for dialogue language understanding}.
\newblock In \emph{Proceedings of the 27th International Conference on
  Computational Linguistics}, pages 1234--1245, Santa Fe, New Mexico, USA.
  Association for Computational Linguistics.

\bibitem[{Huang et~al.(2020)Huang, Feng, Hu, Wu, Du, and
  Ma}]{Huang2020Meta-ReinforcedSystems}
Yi~Huang, Junlan Feng, Min Hu, Xiaoting Wu, Xiaoyu Du, and Shuo Ma. 2020.
\newblock \href {https://www.aclweb.org/anthology/2020.acl-main.636}
  {{Meta-Reinforced Multi-Domain State Generator for Dialogue Systems}}.
\newblock \emph{Proceedings of the 58th Annual Meeting of the Association for
  Computational Linguistics}, pages 7109--7118.

\bibitem[{Jalalvand et~al.(2018)Jalalvand, Ljolje, and
  Bangalore}]{DBLP:journals/corr/abs-1810-00670}
Shahab Jalalvand, Andrej Ljolje, and Srinivas Bangalore. 2018.
\newblock \href {http://arxiv.org/abs/1810.00670} {Automatic data expansion for
  customer-care spoken language understanding}.
\newblock \emph{CoRR}, abs/1810.00670.

\bibitem[{Kobayashi(2018)}]{kobayashi-2018-contextual}
Sosuke Kobayashi. 2018.
\newblock \href {https://doi.org/10.18653/v1/N18-2072} {Contextual
  augmentation: Data augmentation by words with paradigmatic relations}.
\newblock In \emph{Proceedings of the 2018 Conference of the North {A}merican
  Chapter of the Association for Computational Linguistics: Human Language
  Technologies, Volume 2 (Short Papers)}, pages 452--457, New Orleans,
  Louisiana. Association for Computational Linguistics.

\bibitem[{Kurata et~al.(2016)Kurata, Xiang, and Zhou}]{Kurata2016LabeledDG}
Gakuto Kurata, Bing Xiang, and Bowen Zhou. 2016.
\newblock Labeled data generation with encoder-decoder lstm for semantic slot
  filling.
\newblock In \emph{INTERSPEECH}.

\bibitem[{Lei et~al.(2018)Lei, Jin, Kan, Ren, He, and
  Yin}]{Lei2018Sequicity:Architectures}
Wenqiang Lei, Xisen Jin, Min-Yen Kan, Zhaochun Ren, Xiangnan He, and Dawei Yin.
  2018.
\newblock \href {https://doi.org/10.18653/v1/P18-1133} {{Sequicity: Simplifying
  Task-oriented Dialogue Systems with Single Sequence-to-Sequence
  Architectures}}.
\newblock In \emph{Proceedings of the 56th Annual Meeting of the Association
  for Computational Linguistics (Volume 1: Long Papers)}, volume~1, pages
  1437--1447, Stroudsburg, PA, USA. Association for Computational Linguistics.

\bibitem[{Li et~al.(2021)Li, Yavuz, Hashimoto, Li, Niu, Rajani, Yan, Zhou, and
  Xiong}]{li2021coco}
Shiyang Li, Semih Yavuz, Kazuma Hashimoto, Jia Li, Tong Niu, Nazneen Rajani,
  Xifeng Yan, Yingbo Zhou, and Caiming Xiong. 2021.
\newblock \href {https://openreview.net/forum?id=eom0IUrF__F} {Coco:
  Controllable counterfactuals for evaluating dialogue state trackers}.
\newblock In \emph{International Conference on Learning Representations}.

\bibitem[{{Liu} et~al.(2020){Liu}, {Wang}, {Xiang}, and {Meng}}]{textDataAug}
P.~{Liu}, X.~{Wang}, C.~{Xiang}, and W.~{Meng}. 2020.
\newblock \href {https://doi.org/10.1109/CCNS50731.2020.00049} {A survey of
  text data augmentation}.
\newblock In \emph{2020 International Conference on Computer Communication and
  Network Security (CCNS)}, pages 191--195.

\bibitem[{Liu et~al.(2019)Liu, Lim, Suhaimi, Tong, Ong, Ng, Lee, Macdonald,
  Ramasamy, Krishnaswamy, Chow, and Chen}]{liu-etal-2019-fast}
Zhengyuan Liu, Hazel Lim, Nur Farah~Ain Suhaimi, Shao~Chuen Tong, Sharon Ong,
  Angela Ng, Sheldon Lee, Michael~R. Macdonald, Savitha Ramasamy, Pavitra
  Krishnaswamy, Wai~Leng Chow, and Nancy~F. Chen. 2019.
\newblock \href {https://doi.org/10.18653/v1/N19-2004} {Fast prototyping a
  dialogue comprehension system for nurse-patient conversations on symptom
  monitoring}.
\newblock In \emph{Proceedings of the 2019 Conference of the North {A}merican
  Chapter of the Association for Computational Linguistics: Human Language
  Technologies, Volume 2 (Industry Papers)}, pages 24--31, Minneapolis,
  Minnesota. Association for Computational Linguistics.

\bibitem[{Liu et~al.(2020)Liu, Winata, Xu, and Fung}]{liu-etal-2020-coach}
Zihan Liu, Genta~Indra Winata, Peng Xu, and Pascale Fung. 2020.
\newblock \href {https://doi.org/10.18653/v1/2020.acl-main.3} {{C}oach: A
  coarse-to-fine approach for cross-domain slot filling}.
\newblock In \emph{Proceedings of the 58th Annual Meeting of the Association
  for Computational Linguistics}, pages 19--25, Online. Association for
  Computational Linguistics.

\bibitem[{Madotto et~al.(2020)Madotto, Liu, Lin, and
  Fung}]{madotto2020language}
Andrea Madotto, Zihan Liu, Zhaojiang Lin, and Pascale Fung. 2020.
\newblock \href {http://arxiv.org/abs/2008.06239} {Language models as few-shot
  learner for task-oriented dialogue systems}.

\bibitem[{Quan and Xiong(2019)}]{Quan2019EffectiveDA}
Jun Quan and Deyi Xiong. 2019.
\newblock Effective data augmentation approaches to end-to-end task-oriented
  dialogue.
\newblock \emph{2019 International Conference on Asian Language Processing
  (IALP)}, pages 47--52.

\bibitem[{Ramadan et~al.(2018)Ramadan, Budzianowski, and
  Ga{\v{s}}i{\'{c}}}]{Ramadan2018Large-ScaleSharing}
Osman Ramadan, Paweł Budzianowski, and Milica Ga{\v{s}}i{\'{c}}. 2018.
\newblock \href {https://doi.org/10.18653/v1/P18-2069} {{Large-Scale
  Multi-Domain Belief Tracking with Knowledge Sharing}}.
\newblock In \emph{Proceedings of the 56th Annual Meeting of the Association
  for Computational Linguistics (Volume 2: Short Papers)}, volume~2, pages
  432--437, Stroudsburg, PA, USA. Association for Computational Linguistics.

\bibitem[{Rastogi et~al.(2017)Rastogi, Hakkani-Tur, and
  Heck}]{Rastogi2017ScalableTracking}
Abhinav Rastogi, Dilek Hakkani-Tur, and Larry Heck. 2017.
\newblock \href {http://arxiv.org/abs/1712.10224} {{Scalable Multi-Domain
  Dialogue State Tracking}}.
\newblock \emph{2017 IEEE Automatic Speech Recognition and Understanding
  Workshop, ASRU 2017 - Proceedings}, 2018-January:561--568.

\bibitem[{Song et~al.(2020)Song, Zang, Su, Wu, Han, and
  Hu}]{DBLP:journals/corr/abs-2002-09634}
Xiaohui Song, Liangjun Zang, Yipeng Su, Xing Wu, Jizhong Han, and Songlin Hu.
  2020.
\newblock \href {http://arxiv.org/abs/2002.09634} {Data augmentation for
  copy-mechanism in dialogue state tracking}.
\newblock \emph{CoRR}, abs/2002.09634.

\bibitem[{Summerville et~al.(2020)Summerville, Hashemi, Ryan, and
  Ferguson}]{summerville-etal-2020-tame}
Adam Summerville, Jordan Hashemi, James Ryan, and William Ferguson. 2020.
\newblock \href {https://doi.org/10.18653/v1/2020.nlp4convai-1.4} {How to tame
  your data: Data augmentation for dialog state tracking}.
\newblock In \emph{Proceedings of the 2nd Workshop on Natural Language
  Processing for Conversational AI}, pages 32--37, Online. Association for
  Computational Linguistics.

\bibitem[{Thomson and Young(2010)}]{Thomson2010BayesianSystems}
Blaise Thomson and Steve Young. 2010.
\newblock \href {https://doi.org/10.1016/j.csl.2009.07.003} {{Bayesian update
  of dialogue state: A POMDP framework for spoken dialogue systems}}.
\newblock \emph{Computer Speech and Language}, 24(4):562--588.

\bibitem[{Wang and Lemon(2013)}]{wang-lemon-2013-simple}
Zhuoran Wang and Oliver Lemon. 2013.
\newblock \href {https://aclanthology.org/W13-4067} {A simple and generic
  belief tracking mechanism for the dialog state tracking challenge: On the
  believability of observed information}.
\newblock In \emph{Proceedings of the {SIGDIAL} 2013 Conference}, pages
  423--432, Metz, France. Association for Computational Linguistics.

\bibitem[{Wei and Zou(2019)}]{wei-zou-2019-eda}
Jason Wei and Kai Zou. 2019.
\newblock \href {https://doi.org/10.18653/v1/D19-1670} {{EDA}: Easy data
  augmentation techniques for boosting performance on text classification
  tasks}.
\newblock In \emph{Proceedings of the 2019 Conference on Empirical Methods in
  Natural Language Processing and the 9th International Joint Conference on
  Natural Language Processing (EMNLP-IJCNLP)}, pages 6382--6388, Hong Kong,
  China. Association for Computational Linguistics.

\bibitem[{Wen et~al.(2017)Wen, Vandyke, Mrk{\v{s}}i{\'c}, Ga{\v{s}}i{\'c},
  Rojas-Barahona, Su, Ultes, and Young}]{wen-etal-2017-network}
Tsung-Hsien Wen, David Vandyke, Nikola Mrk{\v{s}}i{\'c}, Milica
  Ga{\v{s}}i{\'c}, Lina~M. Rojas-Barahona, Pei-Hao Su, Stefan Ultes, and Steve
  Young. 2017.
\newblock \href {https://aclanthology.org/E17-1042} {A network-based end-to-end
  trainable task-oriented dialogue system}.
\newblock In \emph{Proceedings of the 15th Conference of the {E}uropean Chapter
  of the Association for Computational Linguistics: Volume 1, Long Papers},
  pages 438--449, Valencia, Spain. Association for Computational Linguistics.

\bibitem[{Williams and Young(2007)}]{Williams2007PartiallySystems}
Jason~D. Williams and Steve Young. 2007.
\newblock \href {https://doi.org/10.1016/j.csl.2006.06.008} {{Partially
  observable Markov decision processes for spoken dialog systems}}.
\newblock \emph{Computer Speech and Language}, 21(2):393--422.

\bibitem[{Wu et~al.(2020{\natexlab{a}})Wu, Hoi, Socher, and
  Xiong}]{wu2020todbert}
Chien-Sheng Wu, Steven Hoi, Richard Socher, and Caiming Xiong.
  2020{\natexlab{a}}.
\newblock \href {http://arxiv.org/abs/2004.06871} {Tod-bert: Pre-trained
  natural language understanding for task-oriented dialogue}.

\bibitem[{Wu et~al.(2020{\natexlab{b}})Wu, Hoi, Socher, and
  Xiong}]{wu-etal-2020-tod}
Chien-Sheng Wu, Steven~C.H. Hoi, Richard Socher, and Caiming Xiong.
  2020{\natexlab{b}}.
\newblock \href {https://doi.org/10.18653/v1/2020.emnlp-main.66} {{TOD}-{BERT}:
  Pre-trained natural language understanding for task-oriented dialogue}.
\newblock In \emph{Proceedings of the 2020 Conference on Empirical Methods in
  Natural Language Processing (EMNLP)}, pages 917--929, Online. Association for
  Computational Linguistics.

\bibitem[{Wu et~al.(2019)Wu, Madotto, Hosseini-Asl, Xiong, Socher, and
  Fung}]{wu-etal-2019-transferable}
Chien-Sheng Wu, Andrea Madotto, Ehsan Hosseini-Asl, Caiming Xiong, Richard
  Socher, and Pascale Fung. 2019.
\newblock \href {https://doi.org/10.18653/v1/P19-1078} {Transferable
  multi-domain state generator for task-oriented dialogue systems}.
\newblock In \emph{Proceedings of the 57th Annual Meeting of the Association
  for Computational Linguistics}, pages 808--819, Florence, Italy. Association
  for Computational Linguistics.

\bibitem[{Xie et~al.(2017)Xie, Wang, Li, L{\'{e}}vy, Nie, Jurafsky, and
  Ng}]{DBLP:conf/iclr/XieWLLNJN17}
Ziang Xie, Sida~I. Wang, Jiwei Li, Daniel L{\'{e}}vy, Aiming Nie, Dan Jurafsky,
  and Andrew~Y. Ng. 2017.
\newblock \href {https://openreview.net/forum?id=H1VyHY9gg} {Data noising as
  smoothing in neural network language models}.
\newblock In \emph{5th International Conference on Learning Representations,
  {ICLR} 2017, Toulon, France, April 24-26, 2017, Conference Track
  Proceedings}. OpenReview.net.

\bibitem[{Yin et~al.(2020)Yin, Shang, Jiang, Chen, and
  Liu}]{DBLP:conf/aaai/YinSJCL20}
Yichun Yin, Lifeng Shang, Xin Jiang, Xiao Chen, and Qun Liu. 2020.
\newblock \href {https://aaai.org/ojs/index.php/AAAI/article/view/6491} {Dialog
  state tracking with reinforced data augmentation}.
\newblock In \emph{The Thirty-Fourth {AAAI} Conference on Artificial
  Intelligence, {AAAI} 2020, The Thirty-Second Innovative Applications of
  Artificial Intelligence Conference, {IAAI} 2020, The Tenth {AAAI} Symposium
  on Educational Advances in Artificial Intelligence, {EAAI} 2020, New York,
  NY, USA, February 7-12, 2020}, pages 9474--9481. {AAAI} Press.

\bibitem[{Yu et~al.(2018)Yu, Dohan, Luong, Zhao, Chen, Norouzi, and
  Le}]{DBLP:journals/corr/abs-1804-09541}
Adams~Wei Yu, David Dohan, Minh{-}Thang Luong, Rui Zhao, Kai Chen, Mohammad
  Norouzi, and Quoc~V. Le. 2018.
\newblock \href {http://arxiv.org/abs/1804.09541} {Qanet: Combining local
  convolution with global self-attention for reading comprehension}.
\newblock \emph{CoRR}, abs/1804.09541.

\bibitem[{Zhong et~al.(2018)Zhong, Xiong, and
  Socher}]{Zhong2018Global-LocallyTracking}
Victor Zhong, Caiming Xiong, and Richard Socher. 2018.
\newblock \href {https://doi.org/10.18653/v1/P18-1135} {{Global-Locally
  Self-Attentive Encoder for Dialogue State Tracking}}.
\newblock In \emph{Proceedings of the 56th Annual Meeting of the Association
  for Computational Linguistics (Volume 1: Long Papers)}, pages 1458--1467,
  Stroudsburg, PA, USA. Association for Computational Linguistics.

\end{thebibliography}
\bibliographystyle{acl_natbib}

\clearpage

\appendix
\section{Sample Synthetic Dialogues}
\label{sec:sample_dials}

\begin{table}[h!]
    \scriptsize
    \begin{minipage}[t]{\textwidth}
    \begin{tabular}[]{|l|p{6cm}|}
                \hline
                Speaker &Turn\\
                \hline
                User & \color{violet} I am coming to town and looking for a good Spanish restaurant to try  near the town center. Do you have any suggestions ?\\
                \hline
                Agent &\color{violet} The Spanish restaurant near center is Wagamama. The address is 36 
                saint Andrews street .\\
                \hline
                User & \color{violet} I'll be needing a table for 4 at 5p.m. this next Monday , can you help ?\\
                \hline
                Agent & \color{cyan}Here is your reference number 6huyuygs is there anything else I can do for you?\\
                \hline
                User & \color{cyan}Yes I need to find a train to Cambridge from Kings Lynn.\\
                \hline
                Agent &\color{cyan} When would you like to travel ?\\
                \hline
                User & \color{cyan}Tuesday morning at 9 am\\
                \hline
                Agent &\color{cyan} train tr3447 leaves Kings Lynn at 09:11. Would you like to reserve a ticket ?\\
                \hline
                User & \color{cyan}Yes please. Make a booking for the same number of people .\\
                \hline
                Agent &\color{cyan} Okay, the booking was successful , the total fee is 58.8 gbp payable at the station . reference number is : fz6dqhqt. Anything else i can help you with?\\
                \hline
                User & \color{cyan}That is all i needed today.\\
                \hline
            \end{tabular}
    \end{minipage}
    
    \bigskip
    
    \begin{minipage}[b]{\textwidth}
            \begin{tabular}[]{|l|p{6cm}|}
                \hline
                Speaker &Turn\\
                \hline
                User & \color{blue} I am looking for a train to Cambridge that departs after 10:45. \\
                \hline
                Agent & \color{blue}  What day would you like to travel? \\
                \hline
                User &  \color{blue}  I'll be travelling on a Wednesday. \\ 
                \hline
                Agent & \color{red} Where would you be departing from? \\
                \hline
                User &  \color{red} I will be leaving Norwich.\\
                \hline
                Agent & \color{red} On what day will you be leaving?\\
                \hline
                User &  \color{red} I will be leaving on Wednesday.\\
                \hline
                Agent & \color{brown} It is 18.88 pounds per ticket. How many tickets would you like to purchase? \\
                \hline
                User &  \color{brown}Just 2 tickets please and thank you.\\
                \hline
                Agent &  \color{brown}Your ticket has been reserved . Total fee is 18.88 gbp, payable at the station. Reference number zm5k0a4q. Can i help you with anything else today?\\
                \hline
                User &\color{brown}Can you book Acorn guest house for 7 people starting on Wednesday for 4 nights, please ?\\
                \hline
                Agent &  \color{brown}Thank you so much for reserving my train ticket. I have the information for Acorn guest house you requested.\\
                &\color{brown} Would you like me to reserve it for you ? \\
                \hline
                User &  \color{brown} Yes. I need the reference number.\\
                \hline
                Agent &  \color{brown}The booking was successful. your reference number is dtkw11yu\\
                \hline
                User &  \color{brown}That is all I needed today, thank you for your help.\\
                \hline
            \end{tabular}
            \end{minipage}
    \caption{Sample synthetic dialogues generated by our framework. (T) Dialogue in \textit{restaurant} domain generated by 2 different original dialogues. (B) Dialogue in \textit{train} domain generated by 3 different original dialogues.}
    \label{tab:sampleDialogues}
\end{table}

We showcase two synthetic dialogues generated with our framework, {\itshape cf.} Table~\ref{tab:sampleDialogues}, by merging templates from 2 and 3 different dialogues respectively. Our framework can generate multi-domain dialogues following the original dialogues' belief state. For instance, the first half of the first dialogue includes slots from the \textit{restaurant} domain, while the second half has slots from the \textit{train} domain. The second dialogue on the other hand combines slots from domains: \textit{train} (from two different dialogues) and \textit{hotel} (from another third dialogue). Although both dialogues seem coherent in shape, the latter has a redundancy where the system request the day information after the user already stated it. This is because of a missing annotation where the train-day slot in the belief state of the third turn is missing. These kinds of annotations are unavoidable but negligible because it recaptures a misunderstanding by the agent which is observed in real dialogues frequently.

\end{document}